\documentclass[sigplan,10pt]{acmart}

\usepackage{adjustbox}
\AtBeginDocument{%
  }

\copyrightyear{2024} 
\acmYear{2024} 
\setcopyright{rightsretained} 
\acmConference[EuroMLSys '24]{The 4th Workshop on Machine Learning and Systems}{April 22, 2024}{Athens, Greece}
\acmBooktitle{The 4th Workshop on Machine Learning and Systems (EuroMLSys '24), April 22, 2024, Athens, Greece}
\acmDOI{XX.XXXX/XXXXXXX.XXXXXXX}
\acmISBN{XXX-X-XXXX-XXXX-X/XX/XX}

\usepackage{array}
\usepackage{booktabs}
\usepackage{multirow}
\usepackage{amsmath}
\usepackage{algorithm}
\usepackage{algpseudocode}
\usepackage{hyperref}
\usepackage{caption}
\usepackage{titlesec}
\titlespacing*{\section}{0pt}{0.1ex}{0.1ex}
\usepackage{colortbl}
\usepackage{pbalance}

\begin{document}

\title{Towards Low-Energy Adaptive Personalization for Resource-Constrained Devices}

\author{Yushan Huang}
\affiliation{%
  \institution{Imperial College London}
  \city{London}
  \country{UK}}
\email{yushan.huang21@imperial.ac.uk}

\author{Josh Millar}
\affiliation{%
  \institution{Imperial College London}
  \city{London}
  \country{UK}}
\email{joshua.millar22@imperial.ac.uk}

\author{Yuxuan Long}
\affiliation{%
  \institution{Imperial College London}
  \city{London}
  \country{UK}}
\email{maverick.long22@imperial.ac.uk}

\author{Yuchen Zhao}
\affiliation{%
  \institution{University of York}
  \city{York}
  \country{UK}}
\email{yuchen.zhao@york.ac.uk}

\author{Hamed Haddadi}
\affiliation{%
  \institution{Imperial College London}
  \city{London}
  \country{UK}}
\email{h.haddadi@imperial.ac.uk}



\begin{abstract}

The personalization of machine learning (ML) models to address data drift is a significant challenge in the context of Internet of Things (IoT) applications. Presently, most approaches focus on fine-tuning either the full base model or its last few layers to adapt to new data, while often neglecting energy costs. However, various types of data drift exist, and fine-tuning the full base model or the last few layers may not result in optimal performance in certain scenarios.
We propose Target Block Fine-Tuning (TBFT), a low-energy adaptive personalization framework designed for resource-constrained devices. We categorize data drift and personalization into three types: input-level, feature-level, and output-level. For each type, we fine-tune different blocks of the model to achieve optimal performance with reduced energy costs. Specifically, input-, feature-, and output-level correspond to fine-tuning the front, middle, and rear blocks of the model. We evaluate TBFT on a ResNet model, three datasets, three different training sizes, and a Raspberry Pi. Compared with the $Block Avg$, where each block is fine-tuned individually and their performance improvements are averaged, TBFT exhibits an improvement in model accuracy by an average of 15.30\% whilst saving 41.57\% energy consumption on average compared with full fine-tuning.
\end{abstract}

\titlespacing*{\section}
{0pt}{.5ex}{.2ex}
\titlespacing*{\subsection}
{0pt}{.5ex}{.2ex}




\maketitle

\vspace{-10pt}
\section{Introduction}
\label{sec:intro}

The deployment of deep neural networks on the Internet of Things (IoT) and mobile devices has attracted attention from both industry and academia \cite{murshed2021machine,zhang2023dbgan, huang2021thermal, gong2020experiences, shi2017combining}. In comparison to traditional cloud-based machine learning (ML), on-device ML is often susceptible to personalization issues \cite{incel2023device,dhar2021survey}, where data drift arises between the source (i.e., training) and target (i.e., testing) domains, resulting in a degradation in local model performance. Additionally, due to resource constraints, ML on IoT/mobile devices imposes more stringent requirements in terms of memory, power, and energy consumption \cite{huang2023poster, zhao2022towards,afzal2022battery}.

Several approaches have been proposed to address these issues. One approach is to enhance the domain generalization ability of the base model on the source data \cite{scholkopf2021toward, cha2021swad}. Another approach is to fine-tune the base model on labeled target data to efficiently improve the performance \cite{mairittha2020improving, chen2023train, lai2024adaptive}. The second approach can outperform domain generalization and unsupervised adaptation approaches in terms of model performance such as accuracy \cite{rosenfeld2022domain, kirichenko2022last}. 

Currently, most fine-tuning approaches involve fine-tuning a full model or its last few layers \cite{wortsman2022robust,wang2019better,huang2024microt}. For example, in the domain of large models, a model's front end is considered as a general feature extractor \cite{oquab2023dinov2}. When facing new tasks, only the classifier, typically a fully connected layer, is fine-tuned. However, various types of data drift and personalization issues exist. Liang et~al.~\cite{liang2023comprehensive} categorize data drift into three types: input-, feature-, and output-level. According to the Independent Causal Mechanisms (ICM) principle \cite{scholkopf2012causal, peters2017elements}, the causal generative processes of system variables consist of autonomous modules that do not inform or influence one another. From this perspective, different data shifts should correspond to local changes in the causal generative process. In other words, for various types of data drift, fine-tuning the full model or the last few layers may not always achieve the optimal model performance. This has inspired us to explore whether fine-tuning the blocks corresponding to the different types of data drift can achieve better model performance. Furthermore, full model fine-tuning costs more energy and resources, while parameter selection methods that incorporate parameter selection into training and fine-tuning also impose additional burdens on resource-constrained devices ~\cite{zaken2021bitfit, ding2023parameter,kwon2023tinytrain}. Therefore, it is necessary to explore a low-energy fine-tuning and personalization method that can ensure model performance while reducing energy consumption.

\textbf{Contributions.} 

We introduce Target Block Fine-Tuning (TBFT), a low-energy adaptive personalization framework designed for resource-constrained devices. Specifically, we categorize data drift to input-, feature-, and output-level type, TBFT fine-tunes the front block(s), middle block(s), and rear block(s) to obtain better model performance. 
The `data drift' means the change in statistical properties of the source data compared to the target data \cite{iwashita2018overview}. 
The `block' means a building unit that may encompass multiple layers such as several convolutional and pooling layers, or a single layer such as a fully connected (FC) layer.
We evaluate TBFT on a ResNet model, three datasets, three training sizes, and a Raspberry Pi. Our initial experimental results show that: (i) Compared with the $Block Avg$, where each block (Block1, Block2, Block3, and FC) is fine-tuned individually and their performance improvements are averaged, TBFT exhibits an improvement in model accuracy by an average of 15.30\%; (ii) Compared with the full model fine-tuning, TBFT can save 41.57\% energy consumption on average.

The source code, models, and links to the public datasets are made available on a GitHub repository \footnote{\url{https://github.com/yushan-huang/AdaptivePersonalization}}.


\begin{figure}[t]
\begin{center}
\centerline{\includegraphics[width=0.38\textwidth]{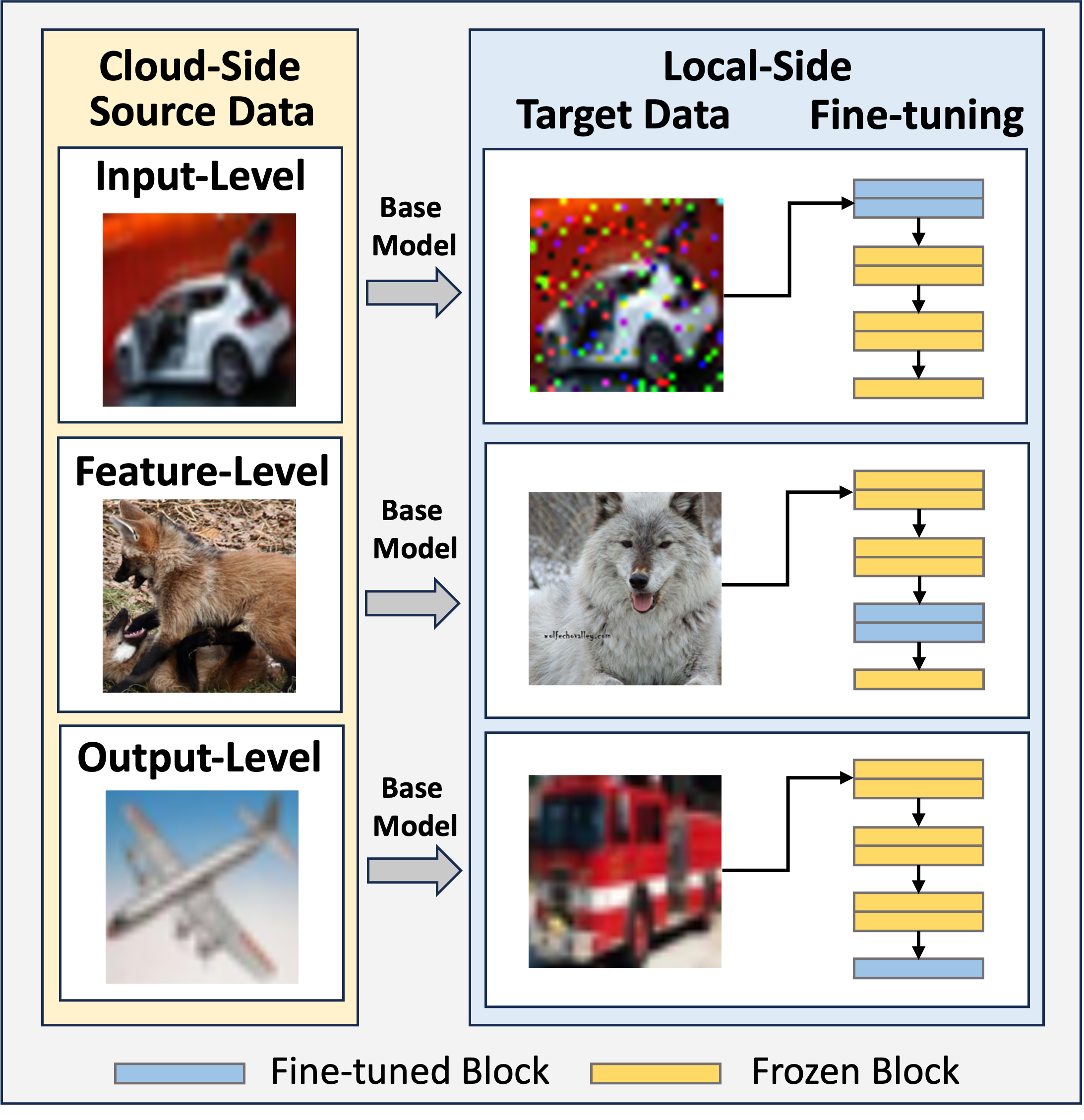}}
\caption{System Overview.}
\label{fig:overview}
\end{center}
\vspace{-20pt}
\end{figure}

\section{Background}
\label{sec:background}



\textbf{Transfer Learning.}
Transfer learning (TL) facilitates the adaptation of pre-trained models to changes in their target domains. Various works have shown that freezing certain layers/parameters during fine-tuning can alleviate overfitting \cite{kirkpatrick, elsa, spottune, catastrophy}, as well as promote more efficient adaptation \cite{autofreeze, flextuning, clint}, resulting in improved TL performance in diverse settings. However, despite extensive efforts to analyze the contributions of model layers towards adaptation for the various types of data drift \cite{neyshabur, equal}, most approaches pre-select fixed layers for all types, usually the last few layers, lacking adaptability at run-time \cite{surgical}. However, TBFT does not fine-tune fixed layers for all types of drift. Instead, it is more flexible and adaptive, fine-tuning different blocks for different types, thereby achieving optimal performance.

\textbf{Personalization.}
One significant issue when deploying ML models in the wild is the personalization of pre-trained features toward shifts in target domain. 
Extensive research has addressed model robustness to real-world domain shifts and adaptation strategies \cite{kumar, andreassen, wiles, innout, salman}. It has been shown that fine-tuning the final layer is often enough for output-level shifts in domains with spurious correlations \cite{kirichenko}. Liang et~al.~\cite{liang2023comprehensive} categorized data drift into input-, feature-, and output-levels, while Lee et~al.~\cite{surgical} found varying sensitivity to these types of drift across different model blocks, prompting us to explore block-specific fine-tuning for optimal performance across drift types. We also explore the approach's stability on small local datasets.

\textbf{Efficient Learning.}
There has been significant work on developing efficient ML models, reducing their size and inference energy cost, for applications on resource-constrained IoT/mobile devices with limited energy availability \cite{tinyml1, tinyml2}. For example, techniques such as pruning, quantization, knowledge distillation, and energy-aware architecture search \cite{nas1, nas2, nas3} have been proposed.
In this paper, we explore efficient learning from another perspective: TBFT only requires adjusting a sub-block of the model to achieve sufficient performance, which reduces the energy cost for personalization. TBFT can also be combined with the various existing efficient learning techniques, such as quantization.

\section{Adaptive Personalization}
\label{sec:system}

\textbf{Assumptions.} We focus on image data and present following scenarios: (i) The base model is trained on the cloud and source datasets, the local datasets are subject to various data drifts with limited quantity. 
(ii) The local devices are sensitive to energy (e.g., energy-harvesting devices) \cite{jeon2023harvnet}.

We categorize data drift into three types: input-, feature-, and output-levels \cite{surgical, liang2023comprehensive}. Input-level drift refers to interference and shifts in pixel-level features in target data compared to source data, such as changes in lighting and noise interference. Feature-level implies that the source and target domains share the same categories but have different sub-populations. Output-level implies differences in output labels between the source and target domains; for example, an image is marked as class 0 in the source data but is marked as class 1 in the target data. We believe that for different types of data drift and personalization, only fine-tuning the corresponding sub-block while freezing the other blocks can achieve sufficient performance. 

In this study, we propose an adaptive block-based fine-tuning approach, TBFT, as shown in Fig. \ref{fig:overview}. On the cloud-side, we obtain a base model on the source data, then on the local-side, TBFT fine-tunes the sub-block(s) corresponding to the drift type to achieve optimal performance. We envision our model $f$, as a composition of function blocks $f_i$, represented as $f(x) = f_n \circ \ldots \circ f_1 (x)$, where each block $f_i$  is equipped with a parameter set $\theta_i$ and may contain several layers. Within the TBFT, we calibrate the parameter set  $\{\theta_i | i \in S\}$ from our selected block $S$, aiming to minimize the loss function $\hat{L}_{tgt}$. The objective is delineated as the optimization problem:

\begin{equation}
\min_{\theta_i, i \in S} \hat{L}_{tgt}(f(\theta_1, \ldots, \theta_n)),
\end{equation}

Here, the set $S$ encompasses the indices of the blocks chosen for fine-tuning. Parameters $\theta_i$ not indexed in $S$ are frozen at their pre-trained values. The block selection depends on the data drift type. For example, we fine-tune the front blocks (e.g., $S = 1$) to capture specific pixel-level variations, or the final block (e.g., $S = n$) to adapt to output-level drift. Consider fine-tuning the first block $S = 1$; it can be illustrated in the reformed model $f^{tgt}$, amalgamating the fine-tuned first block $f^{tgt}_1$ with the subsequent original blocks \cite{surgical}, that:

\begin{equation}
f^{tgt}(x) = f_n \circ \ldots \circ f_2 \circ f^{tgt}_1 (x).
\end{equation}

From a mathematical and logical reasoning perspective, for input-level drift, perturbations such as changes in lighting can be mitigated by fine-tuning the front block. These perturbations are akin to mapping the original input data $x_{src}$ to a new space, producing the perturbed data $x_{trg} = A x_{src}$ through a matrix $A$. This strategy is based on the intuition that the front blocks, which directly interact with the input data, are best suited to adapt to these perturbations. Fine-tuning the front blocks is essentially to directly fit and mitigate these perturbations, whereas adjusting the middle and rear blocks might not achieve optimal performance due to potential information loss in the front block processing.

Similarly, for output-level drift, where there is a change in the target data labels, fine-tuning the rear blocks to mitigate label perturbations can correct the model's predictions. Feature-level drift involves changes in the deep feature distributions of the data, where class labels remain the same, and the data distributions for the same labels in source and target domains may exhibit significant differences. Intuitively, feature-level can be viewed as the case between input- and output-levels, representing a deeper level of input-level perturbation that results in substantial feature distribution differences between the source and target domains, sufficient to impact the model's performance. Therefore, for feature-level, fine-tuning the parameter sets of the middle blocks may compensate for the differences in feature representation between the source and target data.

\begin{figure}[t]
\begin{center}
\centerline{\includegraphics[width=0.35\textwidth]{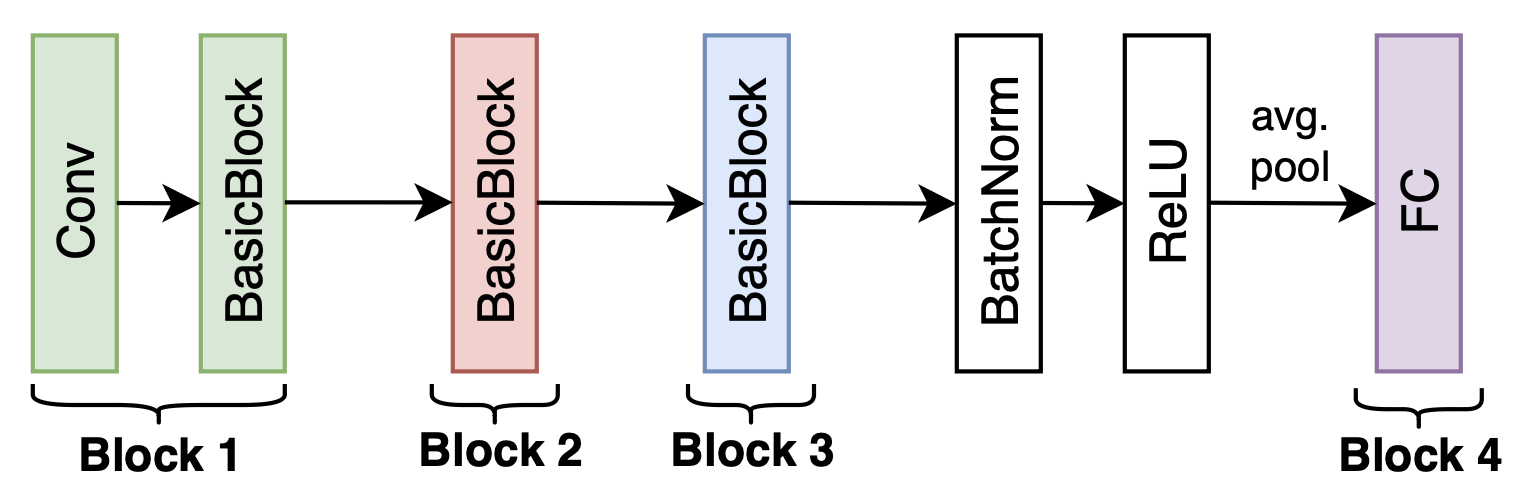}}
\caption{The ResNet-26 model used for all experiments. 
}
\label{fig:resnet_arch}
\end{center}
\vspace{-30pt}
\end{figure}

\section{Implementation \& Evaluation Setup}
\label{sec:implementation}

In this section, we first introduce the experimental model and datasets used (Section ~\ref{sec:modelanddataset}), then detail our implementation and performance metrics (Section ~\ref{sec:prototype}).

\subsection{Model and Datasets}
\label{sec:modelanddataset}

We employ the ResNet-26 \cite{zhang2022memo} model, as shown in Fig. \ref{fig:resnet_arch}. We select this model due to its inherent residual block structure, enabling testing with TBFT without the need for architectural adaptation. 
For each experiment, we train the model from scratch on the source data. We use the Adam \cite{adam} optimizer and early-stop based on accuracy on a held-out validation set. When fine-tuning on the target data, we set a learning rate of  0.01 for the FC layer and 0.001 for the other blocks. The remaining hyperparameters vary depending on the specific experiment (details can be found in Section. \ref{sec:modelperformance}).

We test TBFT on three datasets, one for each category of shift (i.e. input-, feature-, and output-level). To simulate having limited local data, we set the training size to 10\%, 20\%, and 30\% of the full dataset. The validation and testing sizes are both set at 10\%.

\textbf{Input-Level: Cifar10/Cifar10-C.} 
The source distribution, drawn from the original Cifar10 dataset, consists of 32x32 color images with 10 classes. Cifar10-C, the target distribution, consists of corrupted images from the same classes at 5 different levels of severity, representing an input-level shift. There are 14 corruption types (such as frost, Gaussian blur, and Gaussian noise).
For our experiments, we train and evaluate on each of the corruption types individually and report the results averaged across all types. We use the data loading setup from \cite{robustbench}, and train/evaluate only on examples of the highest severity (i.e., corruption level 5).

\textbf{Feature-Level: Living17.} 
This dataset was created as part of the BREEDS benchmark for sub-population shift \cite{santurkar2020breeds}. In BREEDS, ImageNet sub-populations are delineated using a WordNet hierarchy and leveraged to generate various subpopulation datasets. Living17 specifically consists of images of living objects across 17 classes, each with 4 subclasses. The source and target distributions consist of different subclasses. This represents a feature-level shift, since the distribution of features varies between the source and target domains. 


\textbf{Output-Level: Cifar10/Cifar-Flip.} 
To represent an output-level shift, we flip Cifar10 class labels so that each label $y$ becomes $9-y$ \cite{surgical}. For example, label 0 is 9, and so on. The source distribution is the original Cifar10 training set.

\subsection{Prototype and Performance Metrics}
\label{sec:prototype}
We evaluate TBFT on a Raspberry Pi 4 Model B with 2GB RAM, which is commonly utilized as a resource-constrained device due to its limited resource profile. The Pi 4 is equipped with a Broadcom BCM2711 processor, which is a quad-core Cortex-A72 (ARM v8) 64-bit SoC operating at 1.5GHz, offering a balance of computational power and energy efficiency. The memory is 2GB of LPDDR4-3200 SDRAM. 

We evaluate the accuracy and efficiency of TBFT on various training sizes. For efficiency, We utilize the Power Monitoring HAT for Raspberry Pi to measure training system costs: (i) Runtime ($s$), derived from the internal clock. (ii) Memory Usage ($GB$), monitored internally. (iii) Energy Cost ($J$), determined by calculating $E=Pt$, where $P$ is power and $t$ is runtime. We also calculate Energy Saving ($ES$) by $ES=(EB-EF)/EF$, where $EB$ and $EF$ are the energy costs of block-based fine-tuning and full model fine-tuning, respectively. Since the experimental datasets include resolutions of 32x32 (Cifar10-C/Cifar-Flip) and 128x128 (Living17), we conduct measurements with these two resolution conditions; Since our experimental datasets have two different output sizes, 10 (Cifar10-C/Cifar-Flip) and 17 (Living17), we also conduct measurements with corresponding output sizes. Due to memory constraints, we set the batch size to 1 when evaluate the system cost on the Raspberry Pi.

\begin{table}[t]
\footnotesize
\caption{Fine-tuning accuracy results on noised blocks. \\The best block-based accuracy is highlighted. 
}
\footnotesize
\begin{tabular}{cccccc}
 & \multicolumn{5}{c}{Added Noise}                                                                                                                               \\ \cline{2-6} 
 &           & Block 1                          &Block 2                          &Block 3                          & FC                                 \\ \cline{2-6} 
 &Block 1 & \cellcolor[HTML]{FFCC67}84.88±0.23 & 69.24±0.40                         & 43.44±0.38                         & 36.30±0.41                         \\
 &Block 2 & 72.38±0.37                         & \cellcolor[HTML]{FFCC67}83.74±0.21 & 70.54±0.32                         & 67.86±0.35                         \\
 &Block 3 & 56.86±0.37                         & 74.95±0.27                         & \cellcolor[HTML]{FFCC67}83.04±0.29 & 83.85±0.39                         \\
 & FC        & 30.58±0.24                         & 46.94±0.23                         & 51.82±0.41                         & \cellcolor[HTML]{FFCC67}84.28±0.33 \\
 & Block Avg    & 61.18±0.30                         & 68.72±0.28                         & 62.21±0.35                         & 68.07±0.37                         \\
\multirow{-6}{*}{\begin{tabular}[c]{@{}c@{}}T\\ u\\ n\\ e\\ d\end{tabular}} & Full & 83.56±0.28 & 82.87±0.34 & 82.14±0.34 & 83.58±0.42 \\ \cline{2-6} 
\end{tabular}
\label{tab:motivationexp}
\end{table}

\begin{table}[t]
\caption{Overall results of all experiments, including accuracy, energy cost ($E$), and energy-saving rate ($ES$) compared to full model fine-tuning. $I-L$, $F-L$, and $O-L$ correspond to input-, feature-, and output-level drift, and $Block Avg$ to the average accuracy of block-based tuning (excluding full model tuning). 
The best accuracy achieved on each drift is highlighted.}
\footnotesize
\begin{tabular}{cccccc|c}
\hline
             & Block1                       & Block2                       & Block3                       & FC                            & Block Avg      & Full   \\ \hline
I-L Acc (\%) & \cellcolor[HTML]{FFCC67}75.20 & \cellcolor[HTML]{FFFFFF}70.98 & \cellcolor[HTML]{FFFFFF}70.62 & \cellcolor[HTML]{FFFFFF}68.74 & 71.40          & 75.09 \\
F-L Acc (\%) & \cellcolor[HTML]{FFFFFF}48.63 & \cellcolor[HTML]{FFFFFF}57.61 & \cellcolor[HTML]{FFCC67}68.31 & \cellcolor[HTML]{FFFFFF}60.06 & 58.65          & 67.31 \\
O-L Acc (\%) & \cellcolor[HTML]{FFFFFF}21.25 & \cellcolor[HTML]{FFFFFF}38.69 & \cellcolor[HTML]{FFFFFF}64.27 & \cellcolor[HTML]{FFCC67}84.74 & 52.29          & 61.11 \\ \hline
E (J)        & 0.898                         & 0.708                         & 0.564                         & 0.325                         & 0.624          & 1.068 \\
ES (\%)      & 15.92                         & 33.71                         & 47.19                         & 69.57                         & \textbf{41.57} & -     \\ \hline
\end{tabular}
\label{tab:overall}
\end{table}

\begin{table}[t]
\footnotesize
\caption{Training accuracy results with various batch sizes (on the Living17 dataset with 10\% training data). }
\begin{tabular}{cccccc}
\hline
Batch Size & Block 1    & Block 2    & Block 3    & FC         & Full        \\ \hline
128        & 47.30 & 55.57 & 65.83 & 59.01 & 64.69 \\
1          & 45.17 & 54.12 & 64.23 & 57.92 & 62.97 \\ \hline
\end{tabular}
\label{tab:batch}
\end{table}






\section{Preliminary Evaluation Results}
\label{sec:result}

In this section, we evaluate TBFT on model performance and system costs.

\subsection{Motivation Experiments}

To validate whether adjusting different model blocks is sufficient for model personalization towards different data drifts, we introduce normally distributed noise to various model blocks, including the last FC layer. This simulates distribution shifts specifically localized to those parameters. We do this on a ResNet-26 model \cite{zhang2022memo}, trained from scratch on Cifar-10 \cite{krizhevsky2009learning}, fine-tuning the different blocks of the network on Cifar-10 target data while freezing all other parameters. 

Table \ref{tab:motivationexp} shows the results of our motivation experiments; these results show that selectively fine-tuning just the noisy block outperforms fine-tuning other blocks and rivals or exceeds full model fine-tuning, with an average accuracy increase of 18.94\% over $Block Avg$. This indicates targeted block fine-tuning can achieve optimal personalization.

\subsection{Overall Performance}
Table. ~\ref{tab:overall} provides a summary of all experimental results, including accuracy (across three types of drift and with multiple training sizes) and system cost (across all training configurations). We find that fine-tuning Block 1, Block 3, and the FC layer, corresponding to input-, feature-, and output-level drifts, can achieve performance approaching or even surpassing full model fine-tuning. Specifically, compared with $Block Avg$, TBFT can improve model accuracy by 15.30\% on average, suggesting a link between different drift types and the most affected model blocks, where fine-tuning the relevant block can markedly improve performance. In addition, compared with full model fine-tuning, TBFT can save 41.53\% energy cost on average.
Table. ~\ref{tab:overall} summarizes the results from all experiments, with further details in the following section.

\subsection{Single-Batch \& Multi-Batch Training}
When training on resource-constrained devices, the batch size is generally set to one \cite{lin2022device}. However, this can also lead to a decrease in the efficiency of experiments. Therefore, at the beginning of our experiments, we conduct both single-batch training (batch size = 1) on the Raspberry Pi and multi-batch training (batch size = 128) on the GPU, and compare their performance to verify if the latter can be used as an approximate method to increase experiment efficiency. All other training settings are kept consistent. Due to the lower efficiency of single-batch training, we conduct tests only on the Living17 \cite{santurkar2020breeds} dataset, selecting only 10\% of the data as training data. The results are shown in the Table. ~\ref{tab:batch}.

We find that single-batch training and multi-batch training have similar accuracy under the same training configuration. This allows us to use the results from multi-batch training on the GPU for evaluation to increase experiment efficiency. Therefore, unless stated otherwise, we default to reporting model performance based on GPU training.

\begin{figure}[t]
\begin{center}
\centerline{\includegraphics[width=0.45\textwidth]{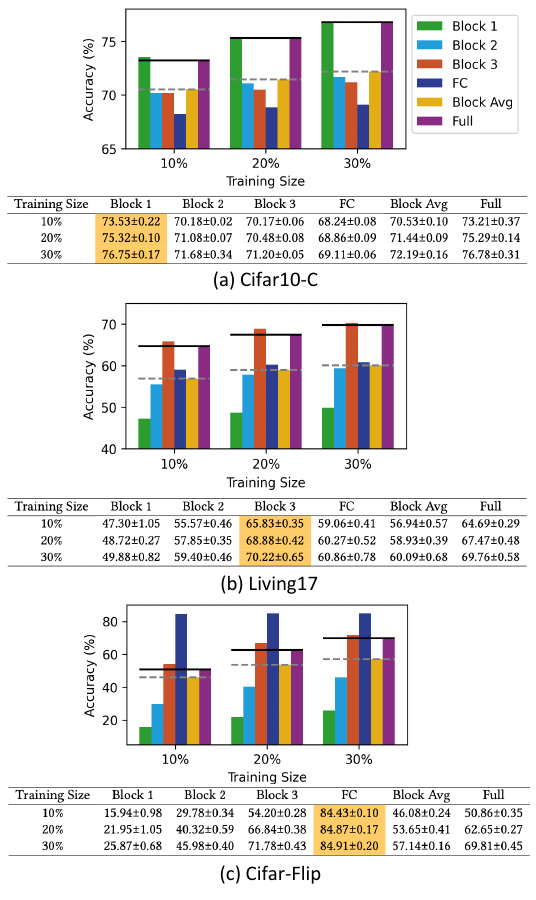}}
\caption{The accuracy results of TBFT on three datasets and with different training sizes. The horizontal solid line shows the accuracy of $Block Avg$, and the horizontal dashed line the accuracy of full model fine-tuning. The highlighted values are the best accuracy across block-based fine-tuning.}

\label{fig:resultall}
\end{center}
\vspace{-20pt}
\end{figure}

\subsection{Model Performance}
\label{sec:modelperformance}

We test the model personalization accuracy for different types of data drift on three datasets: Cifar10-C \cite{hendrycks2019robustness} (input-level), Living17 \cite{santurkar2020breeds} (feature-level), and Cifar-Flip \cite{sagawa2019distributionally} (output-level), using the ResNet-26 \cite{zhang2022memo}. 
To evaluate the accuracy of the proposed method on small local datasets, we test with 10\%, 20\%, and 30\% training sizes.

\textbf{Input-Level: Cifar10-C.} We load the ResNet-26 \cite{zhang2022memo} model and train from scratch on the Cifar10 dataset (with 10 output classes and image resolution of 32x32). Subsequently, we conduct block fine-tuning on the Cifar10-C dataset, activating only one block and freezing the others. The results are shown in Fig. ~\ref{fig:resultall} (a). We find that when processing input-level drift, fine-tuning only Block 1 achieves the best overall model performance, approaching or even surpassing full model fine-tuning. Specifically, compared to $Block Avg$, fine-tuning only Block 1 at training sizes of 10\%, 20\%, and 30\% results in average improvements in model accuracy of 3.00\%, 3.88\%, and 4.56\%, respectively. Moreover, the closer a block is to the front of the model, the better its performance is. The best model performance is achieved by fine-tuning the front-most Block 1 while the worst performance is from fine-tuning the FC layer. Fine-tuning the middle blocks (Blocks 2 and 3) results in moderate performance. These experimental results align with our analysis in Section. \ref{sec:system}, indicating that for input-level drift, the front block of the model is more sensitive than the other blocks.

\begin{table*}[t]
\caption{The time and energy costs of block-based and full model fine-tuning. The $Energy-Saving Rate$ is calculated by comparing the current energy cost to the energy cost of full model fine-tuning.}
\footnotesize
\begin{tabular}{ccccccc|cccccc}
\hline
                & \multicolumn{6}{c|}{Resolution 32, Output 10} & \multicolumn{6}{c}{Resolution 128, Output 17} \\
              & Block 1 & Block 2 & Block 3 & FC    & Block Avg      & All & Block 1 & Block 2 & Block 3 & FC    & Block Avg      & All \\ \hline
Time (s)        & 0.45  & 0.33  & 0.29  & 0.17  & 0.31  & 0.53  & 6.82  & 5.40  & 4.28  & 2.46  & 4.74  & 8.12  \\
Energy Cost (J) & 0.090 & 0.066 & 0.058 & 0.034 & 0.062 & 0.106 & 1.705  & 1.350  & 1.070  & 0.615  & 1.185  & 2.030  \\
Energy-Saving Rate (\%) & 15.09   & 37.74   & 45.28   & 67.92 & 41.51 & -   & 16.01   & 33.50   & 47.29   & 69.70 & 41.63 & -   \\ \hline
\end{tabular}
\label{tab:systemcost}
\end{table*}

\textbf{Feature-Level: Living17.} We train our ResNet-26 base model from scratch on the source dataset from Living17 (with 17 output classes and image resolution of 128x128). Our data preprocessing and augmentation procedures remain consistent with those used in BREEDS \cite{santurkar2020breeds}, which introduced the Living17 dataset. The results are shown in Fig. ~\ref{fig:resultall} (b). We find that for feature-level drift, fine-tuning Block 3 results in the best overall performance, slightly outperforming full model fine-tuning. Compared to $Block Avg$, fine-tuning Block 3 at training sizes of 10\%, 20\%, and 30\% results in average accuracy improvements of 8.89\%, 9.95\%, and 10.13\%, respectively. Moreover, the results indicate that the closer the block to the middle of model, the better the model performance is. For instance, at a training size of 30\%, with Block 3 serving as a reference point (achieving an accuracy of 70.22\%), fine-tuning the adjacent blocks (i.e., Block 2 and the FC layer) results in model accuracy of 59.40\% and 60.86\%, respectively. However, fine-tuning Block 1, which is the farthest from Block 3, yields an accuracy of only 49.88\%. These findings underscore the critical role of the intermediate blocks in capturing and adapting to feature-level drift.

\textbf{Output-Level: Cifar-Flip.} The model setup, training, and fine-tuning procedures are the same as in the Cifar10-C experiments, with the only difference being Cifar10-C is replaced with Cifar-Flip. The results are shown in Fig. ~\ref{fig:resultall} (c). We find that, similar to the previous experimental results, for output-level drift, fine-tuning the blocks closer to the model's rear results in better accuracy, with the best accuracy being achieved through fine-tuning the FC layer. Specifically, compared to $Block Avg$, fine-tuning the FC layer at training sizes of 10\%, 20\%, and 30\% results in accuracy improvements of 38.35\%, 31.22\%, and 27.77\%, respectively. This surpasses fine-tuning the full model, which shows accuracy improvements of 33.57\%, 22.22\%, and 15.10\%, respectively. We also find that fine-tuning blocks closer to the model's rear results in higher model accuracy. For instance, at a training size of 30\%, fine-tuning the block from the front to back results in accuracies of 25.87\%, 45.98\%, 71.78\%, and 84.91\%. Additionally, we can also observe that fine-tuning only the FC layer requires only 10\% training size to achieve performance close to that with 30\% training size. This indicates that for output-level data drift, it is possible to fine-tune the FC layer with a smaller local data size and achieve sufficient model performance.

\textbf{Summary.} Our results indicate that the best model block to fine-tune varies for different types of data drift. Specifically, for input-, feature-, and output-level, the best blocks to fine-tune are the front block (e.g., Block 1), the middle block(s) (e.g., Block 3), and the rear block (e.g., the FC layer). For the scenario where one aspect of the distribution changes while others remain unchanged, it is only necessary to fine-tune the corresponding block to achieve sufficient performance. For input-level drift (e.g., image interference), the pixel-level features of the image change, but the basic structure of the data remains the same between the source and target data. Therefore, it is only necessary to fine-tune the front block to adapt to this image interference. For output-level drift (e.g., label shift), the pixel-level features of the image do not change, and only the output labels have drifted, thus it is only necessary to fine-tune the rear block. Feature-level drift can be considered as somewhere between input- and output-level, where not only is there a significant change in image distribution (i.e. image interference), but this change also brings about an output-level disturbance. Therefore, it is best to fine-tune the middle block(s) of the model. These preliminary results demonstrate that TBFT is a viable way for low-energy personalization in resource-constrained devices.

\subsection{System Cost}

We measure the system cost on a Raspberry Pi by Power Monitoring HAT. The input voltage is 5$V$ and standby memory usage is 0.75$GB$. In both TBFT and full training, with resolution 32 and output size 10, the operational memory usage remains at approximately 0.11$GB$. Similarly, under resolution 128 and output size 17, the operational memory usage is about 0.28$GB$. The memory usage meets the requirements of the experimental device.



The standby power consumption is about 4$W$. The operational power consumption is about 4.20$W$ with resolution 32x32 and output size 10, and about 4.25$W$ with resolution 128x128 and output size 17. The energy cost and time under different configurations are shown in Table. ~\ref{tab:systemcost}. We find that TBFT results in varying degrees of energy savings. Compared to full model fine-tuning, TBFT can achieve a minimum energy saving of 15.09\%, a maximum of 69.70\%, and an average of 41.57\%. Furthermore, across different training configurations, TBFT consistently demonstrates stable energy savings. Overall, analyzing the experimental results in conjunction with model performance and system cost, we find that TBFT can achieve better model performance at lower energy costs.


\section{Discussion and Future Work}
\label{sec:discussion}

\textbf{Key Findings.} 
Our preliminary results indicate that fine-tuning the blocks corresponding to drift types, i.e., TBFT, can achieve better model performance and significantly reduce power consumption. This lays the foundation for achieving low-power personalization on resource-constrained devices. Future work will focus on the following aspects:

\textbf{Discrimination Head.} 
Currently, TBFT relies on prior knowledge of the type of data drift. However, in practical applications, the type of drift is often difficult to know in advance. Some parameter/block selection methods are based on the gradient changes during model training, such as Relative Gradient Norm (Auto-RGN) and Signal-to-Noise Ratio (Auto-SNR) \cite{surgical}. However, these methods embed the parameter/module selection process into the entire training process, which is an additional burden for resource-constrained devices and may affect performance. We plan to approach parameter/block selection from the perspective of the data itself, without embedding it into the training process. For example, we could use information theory \cite{du2014detecting,liang2016personalized} or contrastive learning \cite{lao2023fedvqa, zhang2023doubly} methods to select the parameters/block that need to be trained before model training based on the characteristics of the dataset.

\textbf{Unsupervised Personalization.} We also plan to explore unsupervised personalization methods. Zhang et al. propose MEMO, offering a novel method for unsupervised learning by minimizing the marginal entropy of average predictions for individual images \cite{zhang2022memo}. Considering the implementation principle of MEMO, it may yield good results on input- and feature-level drift, However, for output-level drift, MEMO's strategy of minimizing marginal entropy may backfire, as this optimization direction may conflict with the true distribution of new labels, thereby reducing model performance. In this case, a clustering-based classifier provides a potential solution \cite{wu2020emo}. The core advantage of this approach lies in its reliance on data feature representations rather than label information, enabling it to remain effective even in cases of label inversion. Additionally, the clustering-based classifier may also be effective for input- and feature-level data drifts.

\textbf{Multidimensional Personalization.} In real applications, the data often faces not isolated but complex and composite types of drift, which may simultaneously be affected by input-, feature-, and output-level drifts. This multidimensional data drift necessitates the development of a personalization framework capable of comprehensively addressing these complex scenarios. One possible solution is to first detect and rank the importance of different types of data drift using the Discrimination Head. Subsequently, considering the resource constraints of IoT devices, we adopt a segmented and joint optimization strategy to gradually adapt to different types of data drift, for example, by leveraging the principles of early exit mechanisms \cite{phuong2019distillation}.

\section{Conclusion}
\label{sec:conclusion}

In this paper, we introduced TBFT, a practical low-energy framework to address the challenges of data drift and model personalization on resource-constrained devices. 
TBFT categorizes data drift into three types: input-, feature-, and output-level. For the different types, fine-tuning their corresponding blocks of the model can achieve the best model performance with reduced energy cost.
Our preliminary experimental results show that TBFT exhibits an improvement in accuracy by an average of 15.30\%, and can save 41.57\% energy consumption on average. TBFT's adaptability, energy efficiency, and performance make it a feasible solution for low-energy adaptive personalization on mobile devices.


\bibliographystyle{unsrt}
\bibliography{sample-base}

\end{document}